\documentclass[10pt,twocolumn,letterpaper]{article}

\usepackage{iccv}
\usepackage{times}
\usepackage{epsfig}
\usepackage{graphicx}
\usepackage{subfigure}
\usepackage{amsmath}
\usepackage{amssymb}
\usepackage{url}
\usepackage{multirow}
\usepackage{booktabs}
\usepackage{bm}
\usepackage{color}

\usepackage[pagebackref=true,breaklinks=true,letterpaper=true,colorlinks,bookmarks=false]{hyperref}

\iccvfinalcopy % *** Uncomment this line for the final submission

\ificcvfinal\fi
\begin{document}

%%%%%%%%% TITLE
\title{Unsupervised Learning of Depth and Deep Representation for Visual Odometry from Monocular Videos in a Metric Space}

\author{Xiaochuan Yin, Chengju Liu\\
Tongji University\\
{\tt\small yinxiaochuan@hotmail.com, liuchengju@tongji.edu.cn}
% \and
% \\
% Tongji University\\
}

\maketitle
%\thispagestyle{empty}

%%%%%%%%% ABSTRACT
\begin{abstract}
For ego-motion estimation, the feature representation of the scenes is crucial. Previous methods indicate that both the low-level and semantic feature-based methods can achieve promising results. Therefore, the incorporation of hierarchical feature representation may benefit from both methods. From this perspective, we propose a novel direct feature odometry framework, named DFO, for depth estimation and hierarchical feature representation learning from monocular videos. By exploiting the metric distance, our framework is able to learn the hierarchical feature representation without supervision. The pose is obtained with a coarse-to-fine approach from high-level to low-level features in enlarged feature maps. The pixel-level attention mask can be self-learned to provide the prior information. In contrast to the previous methods, our proposed method calculates the camera motion with a direct method rather than regressing the ego-motion from the pose network. With this approach, the consistency of the scale factor of translation can be constrained. Additionally, the proposed method is thus compatible with the traditional SLAM pipeline. Experiments on the KITTI dataset demonstrate the effectiveness of our method.
\end{abstract}

%%%%%%%%% BODY TEXT
\section{Introduction}
Inferring the structure of scenes and ego-motion through motion is a basic capability of humans. It is also one of the essential research topics in robotics fields, known as the simultaneous localization and mapping (SLAM) problem. This module has become one of the fundamental building blocks for many emerging technologies – from autonomous cars and unmanned aerial vehicles (UAVs) to virtual and augmented reality.

\begin{figure}[t]
\begin{center}
\includegraphics[width=0.75\linewidth, height=0.6\linewidth]{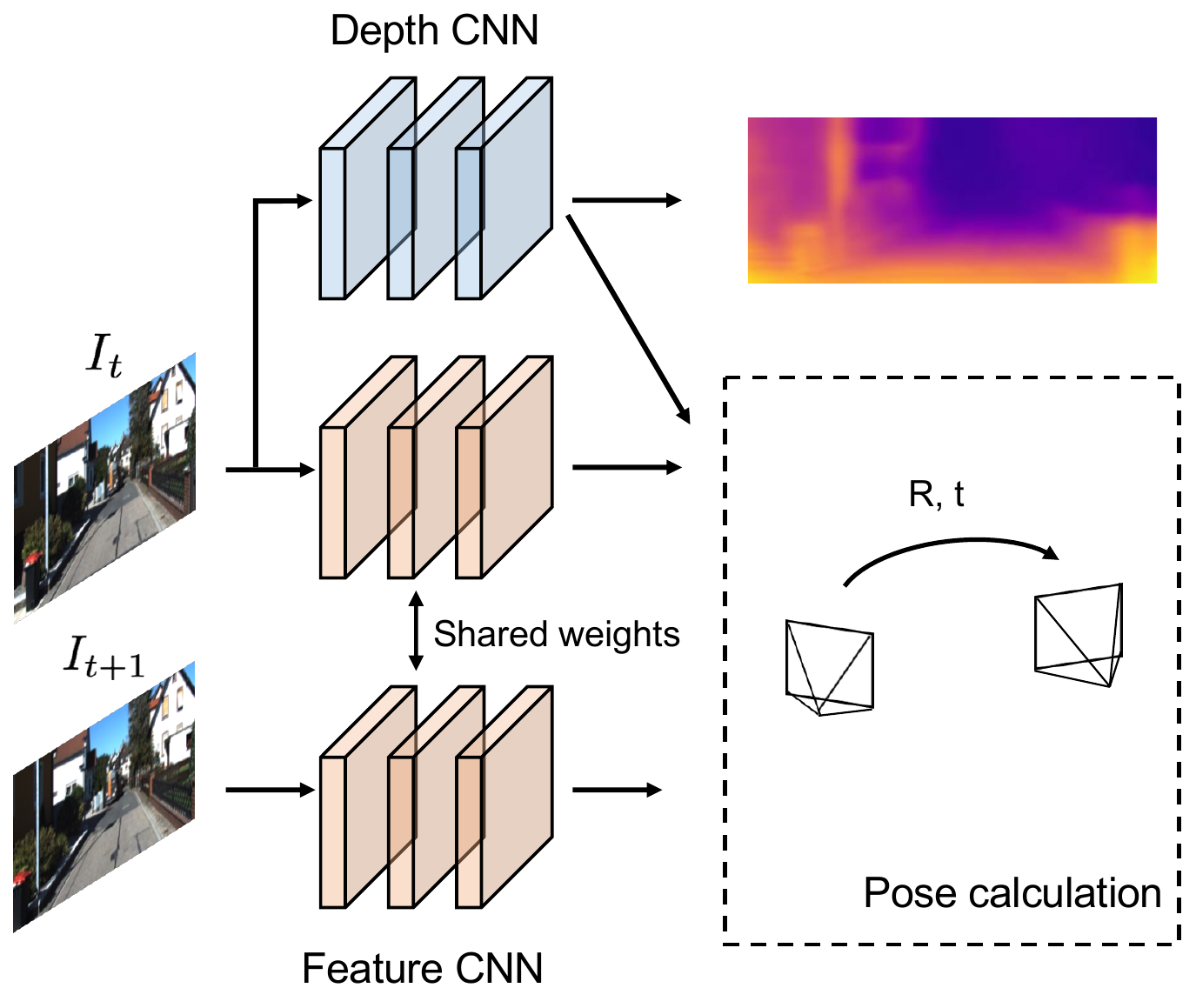}
\end{center}
   \caption{Pipeline of our method for pose and depth estimation. Our method consists of two subnetworks, one for depth prediction and the other for feature representation generation. The camera pose is calculated using the feature and depth pyramids with the direct method.}
\label{fig:scheme_inference}
\end{figure}
% , height=0.8\linewidth
% The SLAM field was dominated by feature-based (indirect) methods for a long time. In recent years, a number of different approaches have gained in popularity such as direct methods and semantic SLAM \cite{bowman2017probabilistic}. Unlike feature-based (indirect) visual odometry methods (e.g. ORB-SLAM \cite{mur2017orb}), direct visual odometry methods use the image intensities information directly to predict the motion and geometry of the scenes. Feature points do not need to be extracted and matched, and the data association and pose estimation are expressed as photometric loss function. The direct visual odometry methods can achieve higher accuracy and robustness for scenes with little feature points. Semantic SLAM, on the other hand, attempts to locate the robots and map the environments with objects or geometry information.

The central idea for a visual odometry method is to calculate the camera motion by locating and aligning the corresponding parts in different viewpoints. Nevertheless, the natural images lie on a nonlinear manifold \cite{weinberger2006unsupervised}, and the nonlinear manifolds of pixel space are locally Euclidean, but globally non-Euclidean. Hence, the selection of valid features and the metric for comparison are crucial for pose estimation. For feature-based methods, camera pose is calculated from the sparse set of keypoints that are extracted and matched between two consecutive images. For direct methods, the ego-motion can be obtained by minimizing the photometric errors of two consecutive images. For learning-based pose estimation methods, especially unsupervised deep learning ones, the poses are regressed from the CNNs directly with the forward movement assumption implicitly (see Section 4.3 for details).
% As a result, both the features and the corresponding metrics need to be learned simultaneously. 

In this work, we present a conceptually simple and unified framework to learn the depth and feature representation from monocular videos. Previous SLAM methods suggest that feature points, structural information and objects can all contribute to the visual odometry problem. On the other hand, visualization of deep neural networks reveals the hierarchical nature of the features in the neural networks: the lower layers of a CNN model capture basic features (e.g., corners, edges) and the higher layers tend to extract high-level task-specific abstractions \cite{zeiler2014visualizing}. Therefore, we construct hierarchical feature pyramids for pose estimation using CNN. We utilize the object-related features to obtain a rough estimate of the ego-motion, and then the pose is refined with low-level local features. The pose is calculated by aligning the features of different viewpoints. A feature selection mask is also obtained to locate the valid features with a self-attention mechanism. Our method can be considered as a unification of different visual odometry methods in the aspect of feature representation. In addition, the ability to abstract and locate features along the multiscale feature pyramid allows the integration with the traditional pipelines, such as  marginalization or keyframe selection. 

% In addition, the spatial encoding of each region of input images can be exploited for further procedures, such as  marginalization or keyframe selection. The ability to abstract the spatial features along the multiscale feature hierarchy allows the integration of pose estimation with the traditional pipelines. 

% The design of DFO has to address two issues: how to compare features for pose estimation without supervised signals and how to abstract valid features at different levels. These two issues are tightly correlated. The first issue means that the features from one point of view can be aligned with features in the new viewpoints. Trival solution should be avoid under no labels. oints of view need to be learned and compared. The second issue is to locate of valid features provides the structural information for pose estimation. 
 % Therefore, non-linear metric learning for image comparison is required. Unlike the previous method, we incorporate the feature code learning and direct visual odometry together. 

In this paper, we propose a novel unsupervised learning framework to learn the feature representation and depth for the visual odometry problem. Specifically, we make the following contributions:
\begin{enumerate}
  \item We can learn the hierarchical feature representation by the metric space for ego-motion estimation. The attention mechanism is applied to provide prior  information at different levels for valid feature selection. 

  \item We present a novel unsupervised pose and depth estimation framework using a direct method without a forward motion assumption.

  % It can incorporate the traditional visual odometry pipeline with deep neural networks in an explicit way.
  % \item We use the attention mechanism to provide masks at different levels of feature pyramids for pose estimation.

  \item We propose to use the 3-frame snippets to maintain the scale consistency of translation and depth. 
  % The ego-motion of camera is calculated using the learned features and estimated depths directly.  
\end{enumerate}

Our unsupervised approach outperforms previous unsupervised methods and achieves comparable results with supervised techniques for the depth estimation task. Compared with other visual odometry methods, the effectiveness of our feature representation learning paradigm are manifested in the pose estimation experiments.

\section{Related Work}
\noindent \textbf{Traditional SLAM methods} Acknowledging the environment's structure through motion is a well-studied problem in the robotics community. The SLAM methods can be mainly divided into three categories: feature-based (indirect) methods, direct methods and learning-based methods.
% For feature-based methods, camera pose is calculated from the sparse set of key-points which are extracted and matched between two input images. For direct methods, the ego-motion can be obtained by minimizing the photometric errors of two consecutive images. These methods may fail under certain conditions such as low texture, stereo ambiguities, occlusions, etc. In order to address these problems, learning-based methods such as deep learning recently have attracted a lot of attention.

The SLAM field was dominated by feature-based (indirect) methods for a long time. In recent years, a number of different approaches have gained in popularity, such as direct methods and semantic SLAM \cite{bowman2017probabilistic}. Unlike feature-based (indirect) visual odometry methods (e.g., ORB-SLAM \cite{mur2017orb}), direct visual odometry methods use the image intensities information directly to predict the motion and geometry of the scenes. Hence, feature points do not need to be extracted and matched, and the data association and pose estimation are expressed as a photometric loss function. The direct visual odometry methods can achieve higher accuracy and robustness for scenes with few feature points. Semantic SLAM, on the other hand, attempts to locate the robots and map the environments with objects or geometric information.

\begin{figure*}[t]
\begin{center}
\includegraphics[width=0.75\linewidth]{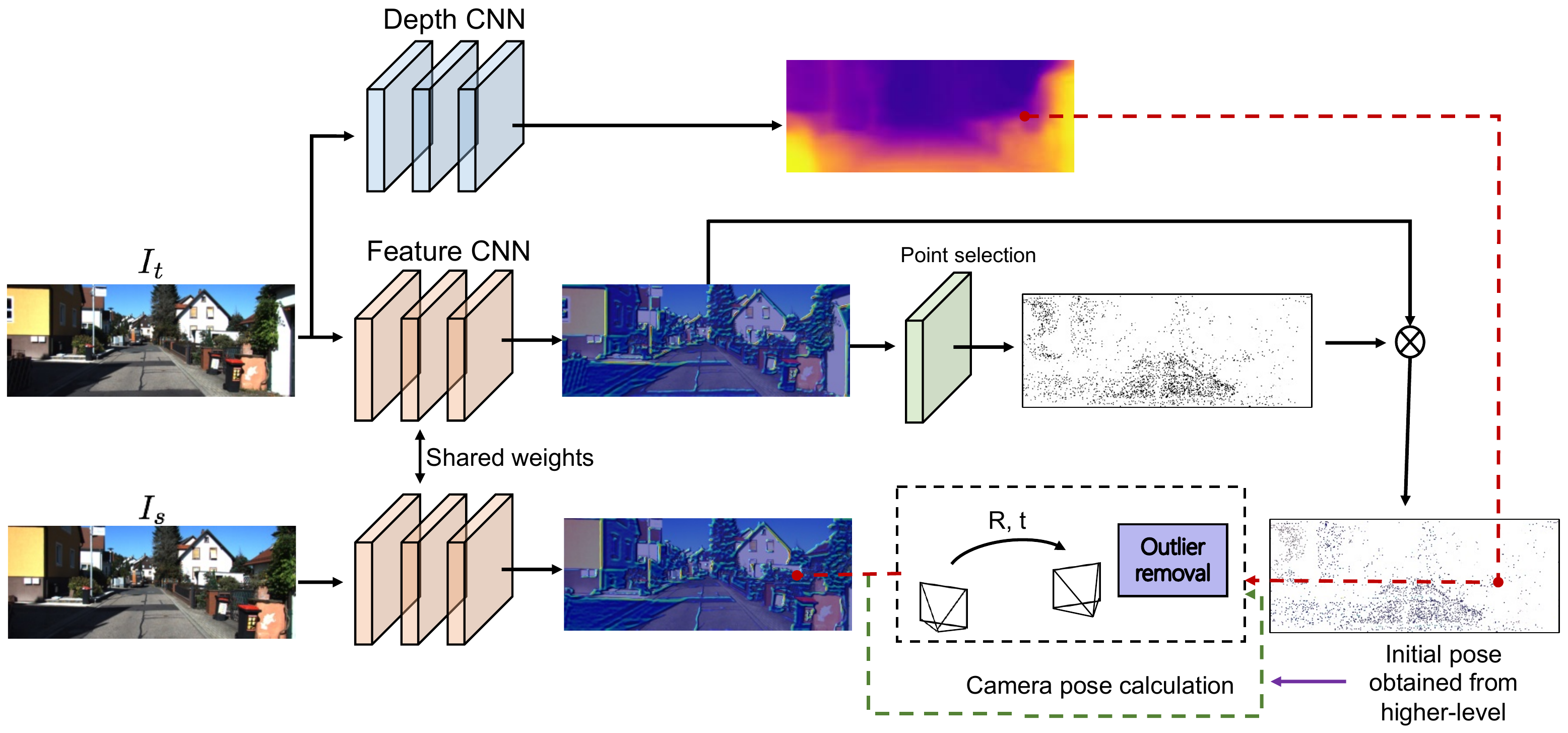}
\end{center}
   \caption{Illustration of our method for the pose calculation process using the $1$-st level depth and feature maps. The initial pose is obtained from the result obtained in the higher level. The feature maps are visualized with Grad-CAM \cite{selvaraju2017grad}, which indicates the portions involved in the calculation of ego-motion.}
\label{fig:scheme_pose_estimation}
\end{figure*}
% , height=0.40\linewidth
% \begin{figure*}[t]
% \begin{center}
% \subfigure[Ours]{
%    \includegraphics[width=0.7\linewidth,height=0.45\linewidth]{pic/scheme2.pdf}
%     \label{fig:scale_1_a}}
% \subfigure[Ours]{
% \includegraphics[width=0.28\linewidth]{pic/ae_net.pdf}\label{fig:scale_1_b}}

% \end{center}
%    \caption{Illustration of the view synthesis constraints of 3-frame snippets. The blue nodes indicate the robot states which contain depth maps for pose perdition.}
% \label{fig:scale_1}

% \end{figure*}

 These methods may fail under certain conditions such as low texture, stereo ambiguities and occlusions. The quality of feature points is essential for the feature-based SLAM method. Agrawal \etal \cite{agrawal2015learning} show that features learned from a convolution neural network outperform the traditional hand-designed features. Learning-based slam may become a promising direction.
 
%  In addition, egomotion is an important source of intrinsic supervision for visual feature learning. Moreover, 3D structure can  be applied as supervision for descriptor learning \cite{schmidt2017self}. 
% Besides feature learning, convolution neural networks have been trained for other tasks for structural understanding, such as depth estimation and pose prediction. 

\noindent \textbf{Unsupervised pose and depth estimation from videos using deep learning} 
Although depth estimation \cite{eigen2014depth,liu2016learning}, pose prediction or combined tasks have made great progress with ground truth structural information collected from sensors such as stereo cameras and LIDAR  \cite{ummenhofer2017demon,yang2018deep}, it is still meaningful to learn the pose and structure of surroundings from monocular videos without supervision. By doing this, large quantities of data can be collected and utilized directly.

Zhou \etal \cite{zhou2017unsupervised} propose a framework that jointly trains the pose estimation and depth prediction networks. In addition to the depth maps and camera poses, the work of Vijayanarasimhan \etal \cite{vijayanarasimhan2017sfm} can predict scene motion from an additional subnetwork. To improve the quality of estimated depth values, several constraints are introduced. In \cite{yin2018geonet}, optical flow estimation is included as an additional module to segment the scenes with rigid structure and nonrigid objects. Mahjourian \etal \cite{mahjourian2018unsupervised}, on the other hand, incorporate the estimated point clouds of two consecutive images as additional constraints. Due to a lack of supervised signals (e.g., the ground truth depth values, poses and stereo images), scale variance of translation and depths may induce vanishing depth problem during training. Wang \etal \cite{wang2018learning} observe this phenomenon and alleviate it by normalizing the inverse depth maps. The depth normalization trick can greatly improve the performance of relative depth estimation at the expense of the scale factor. 

Currently, determining the approach to incorporate the traditional SLAM methods with deep neural networks is still an open problem. The depth prediction neural network can be used in a variety of scenarios, such as keyframe initialization \cite{tateno2017cnn}, scale recovery and virtual stereo constraints \cite{yang2018deep}. CodeSLAM \cite{bloesch2018codeslam} has been proposed as a framework for dense representation of scene geometry using a variational autoencoder. The pose network, however, cannot tightly integrate with the widely applied modules in traditional visual odometry methods (e.g., bundle adjustment, keyframe selection and loop closure) because it cannot locate the corresponding features in the feature maps.
% In \cite{wang2018learning}, the pose network and direct visual odometry are cascaded for pose and depth prediction.
% Our method can learn and locate the valide feature representation for pose estimation using direct method. More importantly, the proposed modules can be plunged into existing direct visual odometry system.

\section{Method}

In this section, we first present an overview of our direct feature odometry method, and then we describe the details of each module and loss function in our algorithm. 

\subsection{Framework Overview}
\begin{figure}[t]
\begin{center}
\includegraphics[width=0.8\linewidth, height=0.6\linewidth]{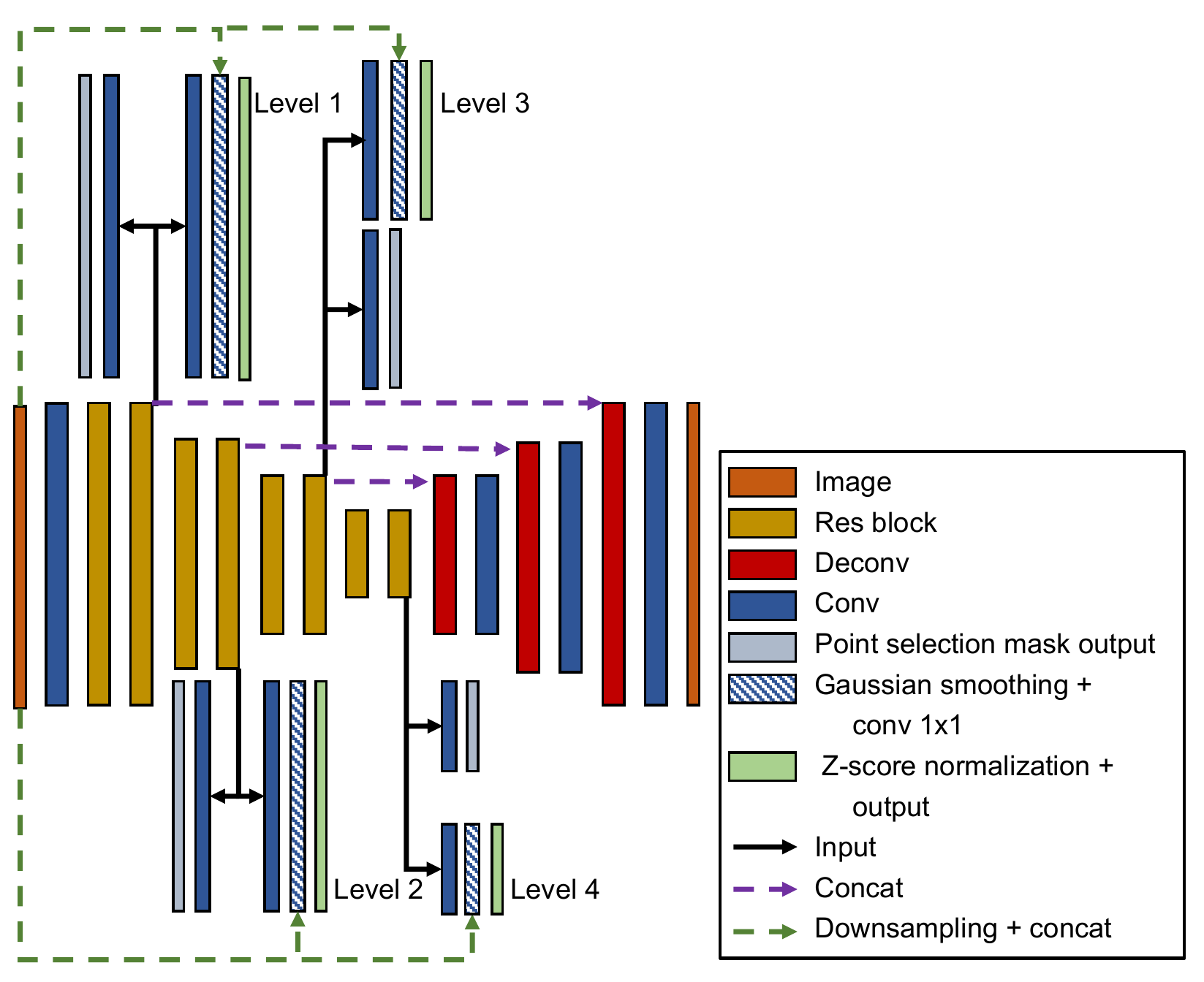}
\end{center}
   \caption{An overview of the deep feature representation network. Given the input image, the multiscale feature pyramid is obtained for pose calculation. We use the U-Net structure to capture the dense feature pyramid using the reconstructed loss function. Spatial information is preserved by concatenating the downsampled images with features. The output feature maps are smoothed using Gaussian smoothing layers. }
\label{fig:ae_net}
\end{figure}

Our framework consists of two subnetworks. One VGG-based network is used to predict the depth pyramid of one input image. The other network aims to produce two feature pyramids of two consecutive input images for pose estimation. The structure of the feature representation network is shown in Figure \ref{fig:ae_net}, and an autoencoder network is used to encode the inputs through nonlinear dimensionality reduction \cite{wang2015deep}.  Due to the ReLU activation in our network, all dimensions in feature space are nonnegative. A Gaussian smoothing kernel (kernel size= $3 \times 3 $) is applied to smooth the feature maps in every channel. 
% In order to get the depth maps and deep feature representation from monocular videos, we use two subnetworks for each task.  The feature maps in different levels for pose calculation are denoted as $\Phi_i, i \in {1, 2, 3,4}$. 

Without loss of generality, we use the pose calculation at the 1-st level to describe the details of our pipeline, which is depicted in Figure \ref{fig:scheme_pose_estimation}. The procedure of our method can be summarized in three main steps:
\begin{enumerate}
  \item We map the consecutive images into the deep feature representation through the feature representation network. Additionally, the feature point selection masks of the target image $\mathbf{W}_t(\mathcal{I}_t)\in \{0,1\}$ are generated from the separate branches  with a self-attention mechanism. The corresponding depth pyramid is obtained through the depth prediction network.
  
  To obtain the feature maps of two views for comparison, a Siamese architecture \cite{chopra2005learning} is applied for the feature generation network, and the shared weights ensure the input images are projected into the same latent space.

  \item We compute the relative pose from multiscale features and depth maps using the direct method. The pose $\bm{\xi}$ is calculated by warping the feature $\widetilde\Phi_s$ and $\Phi_t$ in the selected positions. The pose is initialized with the results from the high-level layer.
  \begin{equation}
    % \bm{\xi} = \arg \min_{\bm{\xi}} \sum \| \phi_s - \phi_t \|_2^2
    \bm{\xi} = \arg \min_{\bm{\xi}} \left \| {\mathbf{W}}_t(\mathcal{I}_t) \odot \left(\widetilde{\Phi}_s(\mathcal{I}_s) - {\Phi_t(\mathcal{I}_t)} \right)\right\|^2
  \end{equation}
  % The second step is to calculate the pose using the feature and depth pyramids. The egomotion of camera are  by aligning the feature maps with estimated depths using direct method. In order to select the valid feature points for pose calculation, feature selection modules are applied as additional branches to generate feature selection masks with self-attention mechanism. 
  
  In each iteration, outliers are removed before calculating the incrementation of the pose. The final egomotion of the camera is obtained through several rounds of iteration.

  \item We warp the source image $\mathcal{I}_s$ to the target image $\mathcal{I}_t$ using the obtained depth and pose to construct the loss function. By minimizing the view synthesis errors between the warped source image $ \widetilde{\mathcal{I}_s}$ and the target image, the parameters of the depth estimation network and feature learning networks, $f_d$ and $f_\phi$, are updated
  \begin{equation}
    f_d, f_\phi = \arg \min_{f_d, f_\phi}  \| \widetilde{\mathcal{I}}_s  - \mathcal{I}_t \|^2.
  \end{equation}

  View synthesis from different points of view as supervision is applied for our unsupervised learning system \cite{zhou2017unsupervised,szeliski1999prediction}. The core idea of this view synthesis pipeline is to reconstruct the target view by sampling pixels from source view given the depth map $\mathbf{Z}$ and relative transformation matrix $\mathbf{T}$. 

  The differentiable bilinear sampling mechanism \cite{jaderberg2015spatial} is used to interpolate the source feature map $\Phi_s$ and source image $\mathcal{I} _ { s }(\mathbf{u} _ { s })$. The homogeneous coordinates of a point in the source view $\mathbf{u} _ { s }$ can be obtained by projecting the point in the target view $\mathbf{u} _ { t }$, which can be formulated as:
  \begin{equation}
    \mathbf{u} _ { s } \sim \mathbf{K} \mathbf{ T } _ { t \to s } \mathbf { Z } _ { t } \left( \mathbf{u} _ { t } \right) {\mathbf{K}}^ { - 1 } \mathbf{u} _ { t }
  \end{equation}
  where $\mathbf{K}$ denotes the camera intrinsics matrix in the corresponding level in the pyramid. 
\end{enumerate}

These three steps demonstrate the whole training pipeline of our method. In the inference stage, we only need the first two steps to obtain the depth maps and camera pose (see Figure \ref{fig:scheme_inference}). In the following, we will describe the details of each module in these three procedures.

% , height=1.0\linewidth

% We are inspired from prior works \cite{zhou2017unsupervised,szeliski1999prediction}, which apply view synthesis as a loss funtion for depth and pose estimation. 

\subsection{Direct Feature Odometry}

Visual odometry is a process to estimate the ego-motion of the camera by minimizing the energy function. The energy function is designed to minimize the difference between the feature point $\mathbf{u}$ in the target feature map $\phi_{t}(\mathbf{u})$ and the corresponding point in the source feature map $\phi_{s}(\omega(\mathbf{u}, d, \bm{\xi}))$. The projective warp function $\omega(\mathbf{u}, d, \bm{\xi})$ projects the point $\mathbf{u}$ in the feature map with the inverse depth $d=z^{-1}$ and transformation $\bm{\xi}$ to the target frame. Lie algebra $\mathfrak{se} (3)$ is chosen as the minimal representation of the camera transformation, which is the tangent space of $SE(3)$ at the identity. The elements of lie algebra $\bm{\xi}$ are mapped to $SE(3)$ by the exponential map $\mathbf{G}=\exp_{ \mathfrak{se} (3)} (\bm{\xi})$ \cite{ma2012invitation}. 
% based on the corresponding inverse depth map $D$ and relative camera transformation $\bm{\xi} \in \mathfrak{se} (3)$. 

% We briefly introduce the relevant mathematical concepts and notation used in our paper. 
% The 3D point $\mathbf{p} = (x, y, z)^T \in \mathbb{R}^3$ is projected to the image coordinates $\mathbf{u}=(u,v)^T \in F$ through the camera projection model $\pi(\mathbf{p})=\mathbf{u}$. The intrinsic camera parameters for projection model are known in advance. During the optimization, 

One of the major contributions of this work is the unsupervised feature representation learning by the metric space. For deep metric learning, the distance of the feature points is obtained by calculating the Euclidean distance in the projected space \cite{hu2014discriminative, andrew2013deep, wang2015deep}.

The distance between two aligned feature maps is:
\begin{equation}
    E(\bm{\xi}) = \sum_i w_t(\mathbf{u}_i) \| \mathbf{r}_i(\bm{\xi},d_i, \mathbf{u}_i)\|_2^2
\end{equation}
where $w_t(\mathbf{u}_i)\in\{0,1\}$ denotes the selection of the feature vector, and $\mathbf{r}_i$ denotes the residual feature vector, which has the following expression:
\begin{equation*}
     \mathbf{r}_i = \phi_{t}(\bm{u}_i) - \phi_{s}(\omega(\mathbf{u}_i, d_i, \bm{\xi}))
\end{equation*}
% $\phi_{\boxplus}(\bm{u}_i)$
where $\phi(\bm{u}_i)$ denotes the small $ K \cdot K $ feature patches around the feature point $\bm{u}_i$. Note that $K$ and $C$ denote kernel and channel sizes, respectively, and $H$ and $W$ denote height and width of feature maps, respectively. This energy function is the maximum likelihood estimate of the rigid body transformation $\bm{\xi}$ between two consecutive frames by assuming i.i.d Gaussian residuals. 

Because there are no labels for feature representation learning, we can not locate the corresponding features in different feature maps directly. Therefore, the alignment of the feature map in each view is performed to minimize the distance in each dimension. We measure the cosine similarity between different viewpoints in each dimension. In order to integrate the similarity measurements with direct method and network training, the features at each level are z-score normalized ($E(\phi_s)=\mathbf{0}$, $E(\phi_t)=\mathbf{0}$, $Var(\phi_s)=\mathbf{1}$ and $Var(\phi_t)=\mathbf{1}$) before comparison. After data normalization, we can calculate the pose by feature comparison because Euclidean distance, Pearson correlation coefficient and cosine similarity are equivalent \cite{berthold2016clustering}. The warping features of two views can be expressed in a compact matrix form as follows:

% The residual of feature code is obtained from errors between two feature points. ($\sum^{H\cdot W}_i\phi_s(\mathbf{u}_i)=\mathbf{0}$, $\sum^{H\cdot W}_i\phi_t(\mathbf{u}_i)=\mathbf{0}$, $\sigma_{\phi_s}=\mathbf{1}$ and $\sigma_{\phi_t}=\mathbf{1}$)

% The warping features of two views can be expressed in a compact matrix form as follows:
\begin{equation}
    \begin{split}
    \min_{\bm{\xi}} &\|\left(\mathbf{w}_t^\mathsf{T}\otimes\mathbf{1}_{(C \cdot K \cdot K) \times 1}\right)\circ \left(\Phi(\mathcal{I}_{t})-\Phi(\mathcal{I}_{s})\mathcal{W}(\mathbf{D},\bm{\xi})\right)\|_F^2 
   % & s.t.  \quad g\left(f(\mathbf{X})\right)=\mathbf{X}
    \end{split}
\end{equation}
where $\mathbf{w}_{t}\in \{0,1\}^{(H \cdot W) \times 1}$ denotes the vectorization of the point selection mask, $\Phi(\mathcal{I}) \in \mathbb{R}^{(C \cdot K \cdot K) \times (H \cdot W)}$ denote the horizontal concatenation of the feature vector $\phi(\mathbf{u})$, $\mathcal{W}(\mathbf{D},\bm{\xi})\in \{0,1\}^{ (H \cdot W) \times (H\cdot W) }$ denotes the warping matrix for feature alignment, and the summation of all the elements in each column of  $\mathcal{W}(\mathbf{D},\bm{\xi})$ is $1$. Additionally, $\otimes$ and $\circ$ denote the Kronecker product and Hadamard product of matrices respectively, $\mathbf{1}_{m\times n}\in \mathbb{R}^{m \times n}$ is a matrix of ones. $\|\cdot\|_F$ denotes the Frobenius norm.

Given the feature representation and depth values, the pose of the camera is solved in an iterative Gauss-Newton procedure. 
% the pose is calculated by aligning the feature maps of different viewpoints.
% It is obtained by warping the features of different view points. Therefore,
% The transformation is calculated with initial value $\bm{\xi}^{(0)}=\mathbf{0}$. 
% $w_i\in\{0,1\}$ denotes the selection of effective points. 
The increment of the pose can be obtained by minimizing the second-order approximation of $ E(\bm{\xi})$:
\begin{equation}
   \sum_i w_{t,i} \mathbf{J}_i^T \mathbf{J}_i \delta \bm{\xi} = - \sum_i w_{t,i} \mathbf{J}_i^T \mathbf{r}_i(\bm{\xi}) 
\end{equation}
where $\mathbf{J}_i:=\nabla r_i(\bm{\xi}^{(0)},\mathbf{u}_i)$ is the Jacobian of the residual with respect to the initial pose $\bm{\xi}^{(0)}$. The updated $\delta \bm{\xi}$ is obtained by using the inverse compositional algorithm \cite{baker2004lucas}, and the Jacobian matrix is computed once per iteration to accelerate the calculation. Accumulated residual errors are decreasing during this process. 

The updated relative transformation is computed as:
\begin{equation}
    % \label{<label>}
    % \mathbf{T}_{i \to j}^{new} = \mathbf{T}_{i \to j} \cdot   \mathbf{G}(\delta \bm{\xi})^{-1}
    \mathbf{T}_{i \to j}^{new} \leftarrow \mathbf{G}(\delta \bm{\xi}) \cdot  \mathbf{T}_{i \to j}
\end{equation}
After several rounds of iteration, we can obtain the final estimated pose.

\noindent \textbf{Point Selection} There are two schemes of point selection: sparse methods and dense methods. Sparse methods use a selected set of points for pose estimation, and dense methods use all pixels with the addition of a geometry prior \cite{engel2017direct}. In this work, a sparse feature mask is learned to select feature points for pose calculation. We apply the pixelwise attention layer to output the per-feature point mask. The probability that selection variable $w_{t,i}$ follows a Bernoulli distribution is $P(w_{t,i}=1) = p$. Label $1$ in the mask indicates that the feature point $i$ in the corresponding location is involved for pose calculation, with label $0$ otherwise. The hard threshold for the feature point selection is nondifferential. Therefore, to achieve the continuous relaxation, the Gumbel Sampling Trick \cite{jang2017categorical, maddison2017concrete, kong2018pixel} is applied to address this issue:
\begin{equation}
  w_{t,i} = softmax \left( \left( \log \left(p)+m \right) \right) /\tau \right)
\end{equation}
where $m$ follows the standard Gumbel distribution and $\tau$ is the temperature parameter. As $\tau \mapsto 0$, the sigmoid function approaches the argmax function. The temperature parameter is initialized to $1$ and decreased to $0.1$ during training, as suggested in \cite{kong2018pixel}.
 
The sparsity of the feature selection mask is constrained by the KL divergence:
\begin{equation}
   \mathcal{L}_{sp} =  KL(\rho||\hat{\rho})=\rho\log(\frac{\rho}{\hat{\rho}}) + (1-\rho)\log(\frac{1-\rho}{1-\hat{\rho}})
\end{equation}
where $\rho$ is the target distribution of feature points, and $\hat{\rho}=\sum_i^{H\cdot W} w_{t,i} / (H\cdot W)$ is the probability of valid points. 
%The sparsity of the feature points can also accelerate the speed of the algorithm for mobile device.

\noindent \textbf{Outlier Removal}. Outliers will worsen the calculation of ego-motion. Therefore, outlier removal is an essential procedure for a visual odometry algorithm. In traditional visual odometry methods, the effects of outliers are reduced by updating the covariance matrix using iteratively reweighted least squares (IRLS). However, in our framework, the deep feature representation is learned rather than using the photometric error of the image's intensity directly. Similar to the DSO method \cite{engel2017direct}, the outliers and occlusion are detected and removed by setting the adapted threshold. If the feature code error surpasses the threshold, the feature points in the corresponding location are removed. The threshold is set as
% \begin{equation}
  $Threshold = \frac{1}{2} \left(median(\|\mathbf{r}\|_2^2) + max(\|\mathbf{r}\|_2^2) \right)$
% \end{equation}
%  $\frac{1}{2} (median(\|\mathbf{r}\|_2^2) + max(\|\mathbf{r}\|_2^2))$ 
 to remove approximately $10\%$ of the feature points.

\begin{figure}[t]
\begin{center}
% \subfigure[Ours]{
%    \includegraphics[width=0.95\linewidth]{pic/scale_1.pdf}
%     \label{fig:scale_1_a}}
% \subfigure[Ours]{
% \includegraphics[width=0.95\linewidth]{pic/scale_2.pdf}\label{fig:scale_1_b}}
\includegraphics[width=0.8\linewidth]{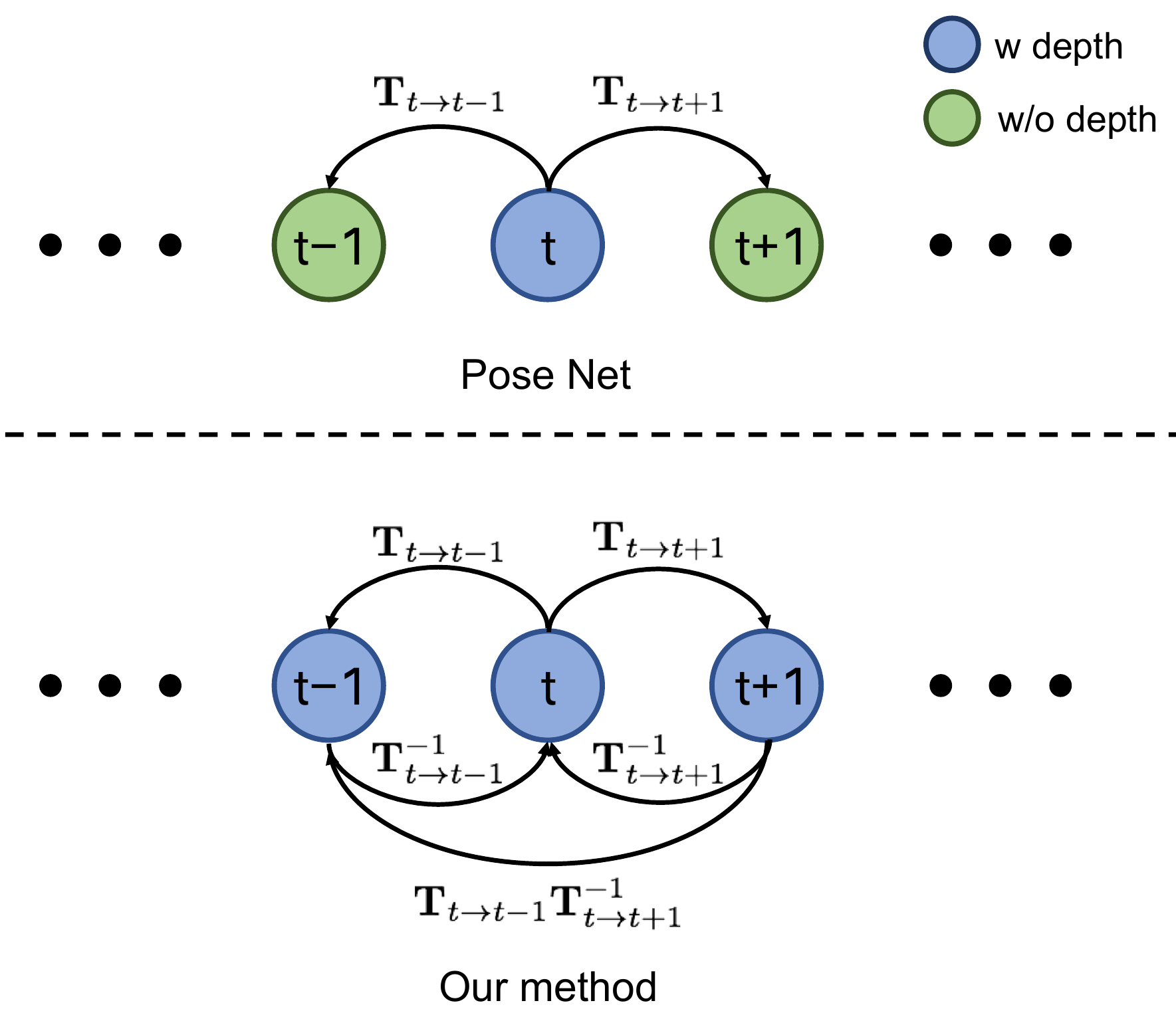}
\end{center}
   \caption{Illustration of the scale consistency constraints using 3-frame snippets. The blue nodes indicate the robot states that contain depth information, while the green nodes do not.}
\label{fig:scale_1}

\end{figure}

\begin{figure}[t]
\begin{center}
\includegraphics[width=0.8\linewidth]{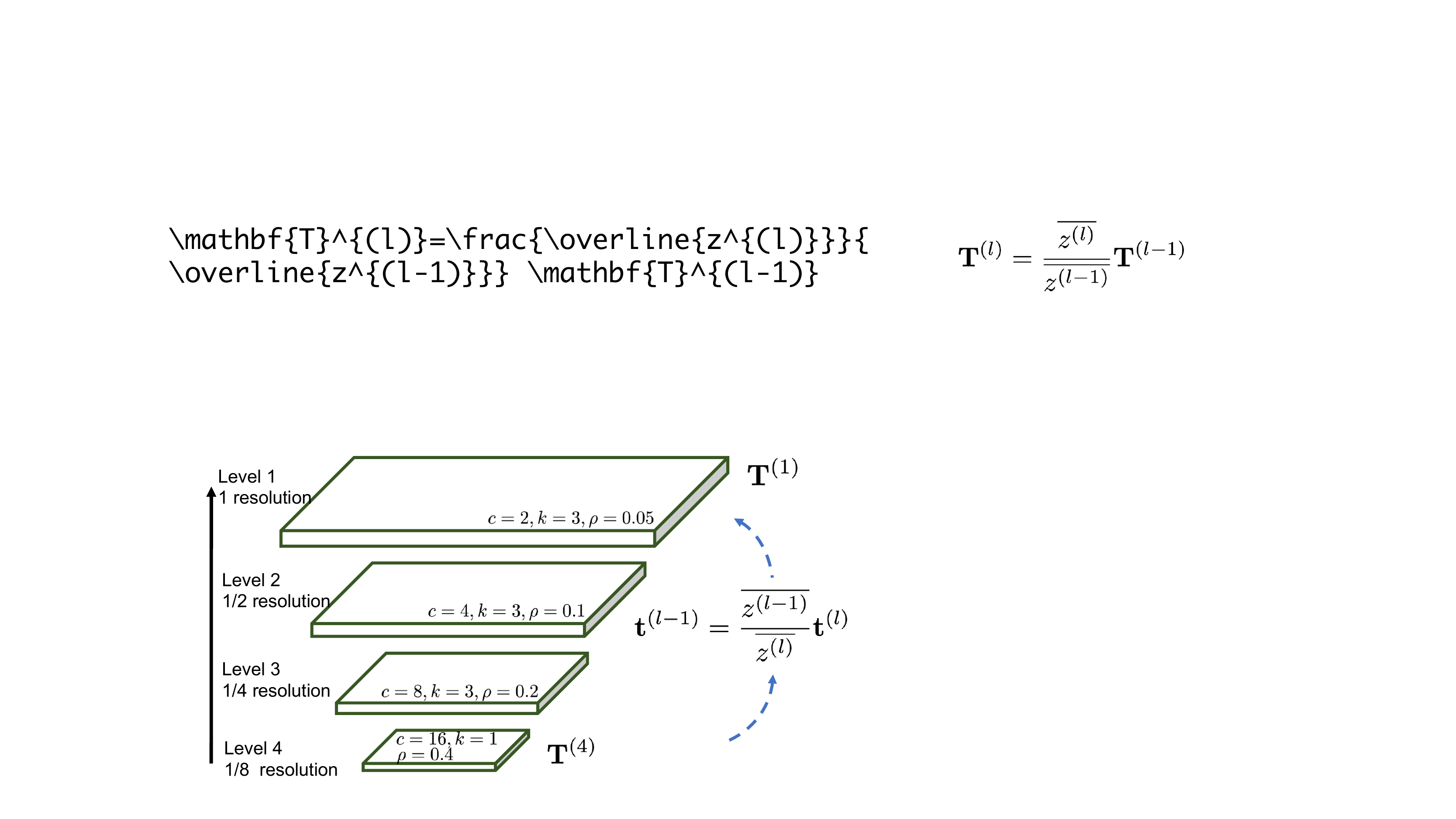}
\end{center}
   \caption{Feature pyramid for pose estimation. The ego-motion is calculated through level 4 to level 1. The initial translation vector of each level is rescaled by multiplying the ratio between the mean depth values  at adjacent levels. c denotes the output channel size, k denotes the patch size for feature comparison and $\rho$ denotes the percentage of the valid feature points involved in the pose calculation. }
\label{fig:scale_2}
\end{figure}

\noindent \textbf{Scale consistency}. The scale of depth values is proportional to that of the translation. Our method can calculate the transformation matrix directly. In this way, the scale factor of translation is affected by the magnitude of depth only. We assume that the depth prediction network can maintain the relative depths through the dataset. The inconsistency of the scale factor comes from two sources: (i) depth scale inconsistency through sequences and (ii) scale inconsistency through the depth pyramid. \\
(i). Pose network regresses the poses and depths directly from two subnetworks which takes three-frame snippets as inputs. The scale factor only depends on the scale of depth in the middle frame. Decreasing the scale of the inverse depth and updating the pose accordingly will lead to the same appearance loss \cite{wang2018learning}. For our method, we replace the pose regression network with the traditional direct visual odometry pipeline. Therefore, the consistency of the depth scale is preserved by using the constrains from three consecutive images. The constraints of depths and poses in 3-frame snippets are shown in Figure \ref{fig:scale_1}. \\
(ii). Scale inconsistency through the depth pyramid (see Figure \ref{fig:scale_2}). For the pose calculation process, both the depth pyramid and the corresponding feature pyramids are used. To ensure the scale consistency for each level, the translation at the $l-1$-th level is initialized with rescaled translation at the $l$-th level in a larger resolution: 
\begin{equation}
    \mathbf{t}^{(l-1)} = \frac{\overline{z^{(l-1)}}}{\overline{z^{(l)}}}\mathbf{t}^{(l)}
\end{equation}
where $\mathbf{t}^{(l)}$ denotes the translation calculated at the $l$-th level, and $\overline{z^{(l)}}$ denotes the mean depth value at the $l$-th depth map. The initial translation at the $4$-th level is set to $\mathbf{t}^{(4)}=\mathbf{0}$, and the rotation matrix is set to the identity matrix $\mathbf{R}^{(4)}=I_3$.
% ,height=0.4\linewidth
\subsection{Training Loss}
\noindent \textbf{Appearance Loss} Similar to previous methods, we use a combination of $L1$ and the structural similarity index (SSIM) \cite{wang2004image} losses to measure the similarity between two images. The $L1$ term captures the low-frequency structure of the images. The SSIM term, on the other hand, maintains the perceptual similarity. The appearance loss is defined as follows:
\begin{equation}
    \mathcal{L}_{ap}=\alpha\frac{1-SSIM(\mathcal{I}_t,  \widetilde{\mathcal{I}_s})}{2} +(1-\alpha)\|\mathcal{I}_t- \widetilde{\mathcal{I}_s}\|_1
\end{equation}
where $\alpha$ is set to $0.85$ and we use $3 \times 3$ filter blocks for the SSIM term. The first term is also known as the structural dissimilarity (DSSIM).

In urban environments, such as in the KITTI dataset, the moving objects and occlusions are considered as outliers. Lowering the weight of the outliers will improve the regression results \cite{belagiannis2015robust}. We directly set the hard threshold to penalize both the DSSIM and  $L1$ error term by: 
\begin{equation}
f(x)=
\begin{cases}
x& x<\epsilon\\
0.1x+0.9\epsilon & x>=\epsilon
\end{cases}
\end{equation}
where $\epsilon$ is the threshold, and $\epsilon$ is set to 0.15 and 0.3 for the DSSIM and $L1$ term, respectively. 

\noindent \textbf{Edge-aware Depth Smoothness Loss}. The smoothness of the depth map is introduced by the $L_1$ penalty on the depth gradients $\partial Z$. The depth gradients are weighted by image gradients because the discontinuities of depth coincide with the edges of input images. 
\begin{equation}
    \mathcal{L}_{smooth} (Z_i)=|\partial_x Z_i| e^{-\|\partial_x \mathcal{I}_i\|} +  |\partial_y Z_i| e^{-\|\partial_y \mathcal{I}_i\|}
\end{equation}
The smoothness constraint of the depth may help the training process. The vanishing disparity is observed in \cite{sfmLearner,wang2018learning} for penalizing the gradients of disparity. 

To summarize, our final objective function becomes:
\begin{equation}
    \begin{split}
    \mathcal{L} = &\sum_l^4 \sum_{i \neq j \in \{1, 2, \ldots, n\}}\mathcal{L}_{ap}(D^{(l)}_i, \mathbf{T}_{i \to j},\mathcal{I}_i^{(l)},\mathcal{I}_j^{(l)})\\
    &+\sum_i^n \frac{\lambda_{sm}}{2^{l-1}}\mathcal{L}_{smooth}(Z_i^{(l)})  +\lambda_{sp} \mathcal{L}_{sp} + \lambda_{ae}\mathcal{L}_{ae}
    \end{split}
\end{equation}
where $\lambda_{sm}$, $\lambda_{sp}$ and  $\lambda_{ae}$ denote the weight of smoothness, sparsity and the autoencoder loss term, respectively, $l$ indicates the scale of the image pyramid, and $n$ is number of images as sequence inputs. The autoencoder loss function $\mathcal{L}_{ae}$ is to ensure the outputs can reconstruct the input images by minimizing the $L2$ error.

\section{Experiments}

In this section, we first describe our network architecture and training details, and then we evaluate the performance of our system on single-view depth and ego-motion prediction tasks. 

\subsection{Implementation Details}
\noindent \textbf{Network Architecture} For the depth estimation task, we apply the same DispNet network as in \cite{mayer2016large,zhou2017unsupervised, wang2018learning,yin2018geonet} for comparison. Given a single view input image, multiscale inverse depth maps can be predicted. We use $1/(10 x + 0.01)$ to obtain the depth map. For the feature representation learning network, the residual attention block \cite{wang2017residual} is used to extract features of different scales and levels for further calculation. The advantage of the residual attention block is that the spatial attention and channel attention are mixed using the bottom-up and top-down feed-forward structure. The spatial information may be lost with the enlarged receptive field in the deeper layers. We concatenate the image intensities together with the deep features to maintain the spatial information. We apply the U-Net structure for feature representation learning. The skip connections in U-Net will accelerate the training process. 

\noindent \textbf{Training Details} Our experiment is conducted using the PyTorch framework \cite{paszke2017automatic}. There are two subnetworks that need to be trained for our method. We adopt a stagewise training strategy. We first train the autoencoder network to obtain the features of different levels. For the depth estimation network, the initialization of this network is crucial. It can be obtained either from \cite{zhou2017unsupervised} or from our methods trained with images only. Then, the two networks are jointly trained in an end-to-end fashion.

During training, we resize the image sequences to a resolution of $128\times416$ and remove the static frames. We also perform random resizing, cropping, and other color augmentations to prevent overfitting during training with a $50\%$ chance. For color augmentations, we perform the random gamma, brightness, and color shifts by sampling from uniform distributions in the range $[0.9, 1.1]$ with a $50\%$ chance. The two subnetworks are initialized with \cite{he2015delving}. The kernel size of the small patch, the channel size of the output features and the probability of selected points at each level of the feature representation learning network are shown in Figure \ref{fig:scale_2}. The network is optimized by Adam \cite{kingma2014adam}, where $\beta_1=0.9, \beta_2 = 0.999$. We use an initial learning rate of $1\times 10^{-5}$ and drop to $1\times 10^{-6}$. The weights of the loss terms are set to $\lambda_{sm}=0.1$, $\lambda_{sp}=0.01$ and $\lambda_{ae}=0.01$. The training process typically takes approximately 30 epochs to converge. 

% In \cite{wang2018learning}, the normalization of disparity trick 
% \begin{equation*}
%     \tilde{d_i} = \frac{N d_i}{\sum_{j=1}^N d_j}
% \end{equation*}
% is applied to solve the scale ambiguity between translation and depth, which is also used for key frame selection in LSD-SLAM \cite{engel2014lsd}. This trick can significant improve the depth estimation results by eliminating the scale of estimated depth map. Therefore, the relative depth estimation networks are obtained. The experimental results of DDVO \cite{wang2018learning} without the depth normalization are shown in Table \ref{tab:kitti_depth}. 
\subsection{Visualization of the Feature Maps}
\begin{figure}[t]
  \begin{center}
  \includegraphics[width=0.98\linewidth]{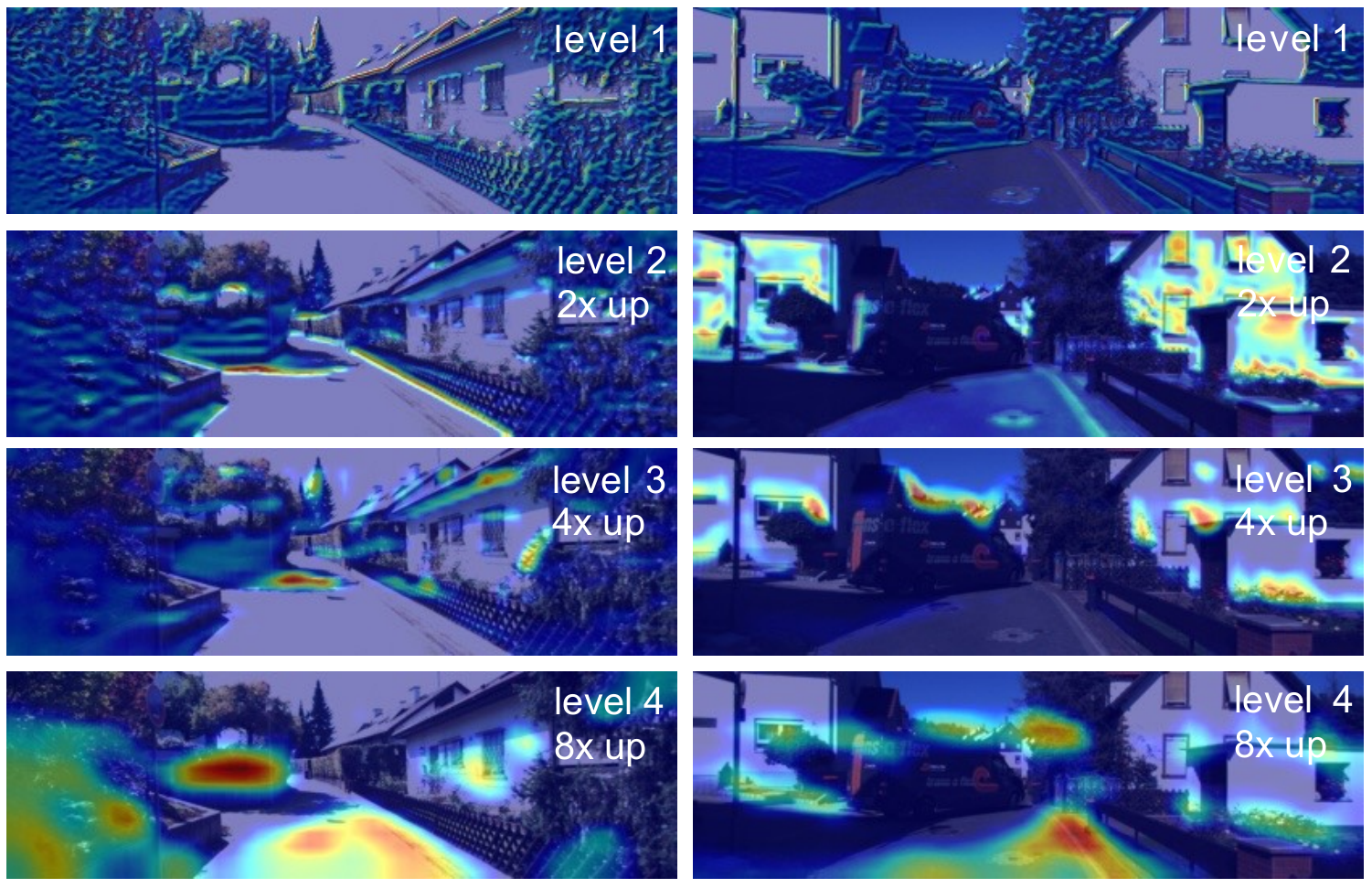}
  \end{center}
     \caption{Visualization of the feature maps involved in the calculation of motion.}
  \label{fig:attention}
  \end{figure}
  
  We construct multiscale feature and depth pyramids for our method. Different resolutions of the feature and depth maps are learned separately. The feature regions involved in the pose estimation are illustrated using Grad-CAM \cite{selvaraju2017grad}. As shown in Figure \ref{fig:attention}, different levels of feature maps can learn the feature representations in a hierarchical way. The pose is calculated through high-level to low-level features in enlarged feature maps.
%   To ensure the consistency of the scale factor between different layers, the translation is updated by multiplying the ratio between the mean depth values at adjacent levels. The rescaled pose values are used for further calculation in a larger resolution. 
\subsection{Monocular Depth Estimation}
We implement the same split of Eigen \etal \cite{eigen2014depth}  to evaluate the performance of our DFO framework with others in the single-view depth estimation task. We use the same VGG-based network and follow the same training protocol as in \cite{zhou2017unsupervised,yin2018geonet,mahjourian2018unsupervised,wang2018learning}. Note that the modification of the neural network architecture and resolution of input images will affect the performance of the algorithm. The length of image sequences is set to 3. The ground truth depth maps are projected from point clouds using LiDAR for evaluation. 

As shown in Table \ref{tab:kitti_depth}, our method performs better than supervised methods (i.e., Eigen \etal \cite{eigen2014depth}, Liu \etal \cite{liu2016learning} and Godard \etal \cite{godard2017unsupervised}) and unsupervised methods (i.e., Yin \etal \cite{yin2018geonet}, Zhou \etal \cite{zhou2017unsupervised} and Mahjourian \etal \cite{mahjourian2018unsupervised}). Our method does not use other constraints such as ICP (i.e., Mahjourian \etal \cite{mahjourian2018unsupervised}) and optical flow (i.e., Yin \etal \cite{yin2018geonet}). Therefore, the performances of our method could be further improved by incorporating these loss functions. 

Our method performs comparably to the results that consider the depth normalization trick ($\tilde{d_i} = \frac{N d_i}{\sum_{j=1}^N d_j}$) in \cite{wang2018learning}. The normalization of the disparity trick is applied to solve the scale ambiguity between translation and depth, which is also used for key frame selection in LSD-SLAM \cite{engel2014lsd}. It can significantly improve the depth estimation results by eliminating the scale of the estimated depth map. Therefore, the relative depth estimation networks are obtained. The experimental results of DDVO \cite{wang2018learning} without the depth normalization are shown in Table \ref{tab:kitti_depth}. The proposed method, on the other hand, can preserve the scale of the depth map. 

We also train our model on the Cityscapes dataset (10 epochs) and then fine tune it on the KITTI dataset (5 epochs). The results show that the supervised methods (i.e., Godard \etal \cite{godard2017unsupervised}) can notably benefit from the enlarged training set. The moving objects that appear in the Cityscapes dataset hamper the further improvement of our method even when more training data are involved.

We perform an ablation study of our method. The results manifest that the learned feature representation can greatly improve the performances. The sparse visual odometry achieved by point selection in the feature map can also benefit the accuracy of the estimated depth values.

% \textcolor{red}{As shown in Table 1, “Ours VGG” trained only on KITTI shares the same network architecture with “Zhou et al. [56] without BN”, which reveals the effectiveness of our loss functions. While the difference between “Ours VGG” and “Ours ResNet” validates the gains achieved by different net- work architectures. Our method significantly outperforms both supervised methods [9, 28] and previously unsuper- vised work [14, 56]. A qualitative comparison has been vi- sualized in Figure 4. Interestingly, our result is slightly inferior to Godard et al. [15] when trained on KITTI and Cityscapes datasets both. We believe this is due to the profound dis- tinctions between training data characteristics, i.e. rectified stereo image pairs and monocular video sequences. Still, the results manifest the geometry understanding ability of our GeoNet, which successfully captures the regularities among different tasks out of videos.}

\begin{table*}[!htbp]
% \small
\footnotesize
% \scriptsize 
\begin{center}
\begin{tabular}{l | c | c | c c c c | c c c }
\toprule
\multirow{2}{*}{Method} & \multirow{2}{*}{Supervised}& \multirow{2}{*}{Dataset}&
\multicolumn{4}{c|}{Error metric}&
\multicolumn{3}{c|}{Accuracy metric} \\
\cline{4-7} \cline{8-10}  &  & & Abs Rel& Sq Rel & RMSE & RMSE log & $\delta  < 1.25$ & $\delta  < 1.25^2$ & $\delta  < 1.25^3$ \\

% \hline
% Method & Supervised & Dataset & Abs Rel& Sq Rel & RMSE & RMSE log & $\delta  < 1.25$ & $\delta  < 1.25^2$ & $\delta  < 1.25^3$\\
% \midrule
\hline

Eigen \etal \cite{eigen2014depth} & Depth & K & 0.203 &  1.548 & 6.307 & 0.282 & 0.702 & 0.890 & 0.958\\
Liu \etal \cite{liu2016learning} & Depth & K & 0.202 &  1.614 & 6.523 & 0.275 & 0.678 & 0.895 & 0.965\\
Godard \etal \cite{godard2017unsupervised} & Pose & K & 0.148 &  1.344 & 5.927 & 0.247 & 0.803 & 0.922 & 0.964\\
Zhou \etal \cite{zhou2017unsupervised} & No & K & 0.208 &  1.768  & 6.856  & 0.283& 0.678 & 0.885 & 0.957\\
Yin \etal \cite{yin2018geonet} VGG & No & K & 0.164 &  1.303  & 6.090 & 0.247 & 0.765 & 0.919 & 0.968\\
Mahjourian \etal \cite{mahjourian2018unsupervised} & No & K & 0.163 &  1.240 & 6.220 & 0.250  & 0.762 & 0.916 & 0.968\\
Wang \etal \cite{wang2018learning} & No & K & 0.213  & 3.770  & 6.925  &  0.294 & 0.758 & 0.909 & 0.958\\
Wang \etal \cite{wang2018learning} DN & No & K & 0.151  & 1.257  & 5.583  &  0.228 & 0.810 & 0.936 & 0.974\\
\midrule
Our method (Image only) & No & K & 0.247  & 2.003  & 7.239  &  0.318 & 0.612 & 0.855 & 0.943 \\
Our method (w/o PS) & No & K & 0.156  & 1.151  & 5.580  &  0.228 & 0.789 & 0.930 & 0.974 \\
Our method & No & K & 0.152  & 1.172  &5.576  &  0.226 & 0.797 & 0.933 & 0.975\\
% \hline
\midrule
Godard \etal \cite{godard2017unsupervised} & Pose & CS+K & 0.124 & 1.076  & 5.311 & 0.219 & 0.847 & 0.942 & 0.973\\
Zhou \etal \cite{zhou2017unsupervised} & No & CS+K & 0.198 &  1.836  & 6.565  & 0.275 & 0.718 & 0.901 & 0.960\\
Mahjourian \etal \cite{mahjourian2018unsupervised} & No & CS+K & 0.159 &  1.231 & 5.912 & 0.243  & 0.784 & 0.923 & 0.970\\
Our method & No & CS+K  & 0.152 & 1.151 &   5.577  &  0.226 & 0.794  &  0.933 & 0.976\\
% Wang \etal \cite{wang2018learning} DN & No & CS+K & 0.148 & 1.187 & 5.496 &  0.226 & 0.812 &0.938 & 0.975\\
% Our method & No & K & 0.152  & 1.172  &5.576  &  0.226 & 0.797 & 0.933 & 0.975\\
\bottomrule
\end{tabular}
\end{center}
\caption{Monocular depth results on KITTI 2015 \cite{menze2015object} with the split of Eigen et al. \cite{eigen2014depth} . For the dataset column, K and CS denote training on the KITTI \cite{Geiger2012CVPR} and Cityscapes datasets \cite{cordts2016cityscapes} respectively. DN is an acronym for depth normalization, and PS represents point selection. The results for other methods are taken from \cite{godard2017unsupervised,yin2018geonet,zhou2017unsupervised,mahjourian2018unsupervised,wang2018learning}. Note that in the results from  Wang \etal \cite{wang2018learning}, DN eliminates the scale of depth maps. }
\label{tab:kitti_depth}
\end{table*}
\subsection{Camera Pose Estimation}

The pose estimation method in our framework is evaluated with the official KITTI odometry split. The dataset contains 11 driving sequences with the ground truth pose captured  with IMU/GPS. We follow the same training and testing split as in \cite{zhou2017unsupervised, yin2018geonet, mahjourian2018unsupervised, wang2018learning}; the 00-08 sequences are used for training and the 09-10 sequences for testing. The ORB-SLAM \cite{mur2017orb} (feature-based  monocular visual odometry method) and the DSO method \cite{engel2017direct} (direct method) are compared with our method. There are two variants of the ORB-SLAM for comparison. The ORB-SLAM (short) version takes 5-frame snippets as inputs. The ORB-SLAM (full) version, on the other hand, takes the entire trajectory as inputs, and the calculated poses are optimized with postprocessing procedures such as loop closure, bundle adjustment and relocalization. The DOS (full) also takes full sequences and postprocessing steps. Our method calculates the ego-motion from feature maps of two input images, and the scale factor of the translation is consistent. We simply integrate the relative pose of two consecutive views for pose evaluation. The pose estimation results are shown in Table \ref{tab:kitti_pose}.

% \textcolor{red}{We have evaluated the performance of our GeoNet on the official KITTI visual odometry split. To compare with Zhou et al. [56], we divide the 11 sequences with groundtruth into two parts: the 00-08 sequences are used for training and the 09-10 sequences for testing. The se- quence length is set to be 5 during training. Moreover, we compare our method with a traditional representative SLAM framework: ORB-SLAM [32]. It involves global optimization steps such as loop closure detection and bun- dle adjustment. Here we present two versions: “The ORB- SLAM (short)” only takes 5 frames as input and “ORB- SLAM (long)” takes the entire sequence as input. All of the results are evaluated in terms of 5-frame trajectories, and scaling factor is optimized to align with groundtruth to resolve scale ambiguity [43]. As shown in Table 3, our method outperforms all of the competing baselines. Note that even though our GeoNet only utlizes limited informa- tion within a rather short sequence, it still achieves better result than “ORB-SLAM(full)”. This reveals again that our geometry anchored GeoNet captures additional high level cues other than sole low level feature correspondences. Fi- nally, we analyse the failure cases and find the network sometimes gets confused about the reference system when large dynamic objects appear nearby in front of the camera, which commonly exist in direct visual SLAM [10].}

As shown in Table \ref{tab:kitti_pose}, our method outperforms other direct visual odometry methods by a large margin even without postprocessing. As shown in Table \ref{tab:kitti_pose}, we compare the results with vary length of inputs. It shows that the accumulated errors will worsen the performance of our method. Our results can be further improved by involving postprocessing steps in the traditional visual odometry pipeline.

\noindent \textbf{Comparison with the pose network} The pose network takes a sequence of images as inputs and outputs the corresponding poses. As shown in Table \ref{tab:kitti_pose}, the estimated poses using our method outperform both the feature-based and the direct visual odometry methods. Our method falls short of the pose regression methods in Seq. 9. The pose regression network implicitly integrate the feature selection, matching for pose estimation with the forward movement assumption. To verify this limitation, we shuffle or duplicate the input images using the pretrained networks from \cite{zhou2017unsupervised,sfmLearner}, the pose estimation network still outputs the forward motion and the speed of the vehicle is relatively stable. The result indicates that only forward movement is learned. The reason of this phenomenon is that there is no constrains between the output poses. We also train the pose network with both forward and backward sequences using \cite{zhou2017unsupervised} (see Table \ref{tab:kitti_pose}), and the results indicate that the pose network cannot differentiate the varieties of motions, while maintaining the similar depth estimation results (see Table \ref{tab:kitti_depth}). No such movement assumption is applicable to our method. Our method can deal with different kinds of movements because the ego-motion of the camera is calculated with the feature selection and comparison in an explicit way. 

% , which is contrary to expectation
% Therefore, date augments and collection should be made to handle different kinds of situations, for example, backward and static situations. , because the speed of vehicle is relative stable in the training set

\begin{table}[!htbp]
% \small
\footnotesize
% \scriptsize 
\begin{center}
\begin{tabular}{l | c | c | c }
\toprule
% \hline
Method & Type & Seq. 09 & Seq. 10\\
% \midrule
\hline
ORB-SLAM (full)& F & 0.014 $\pm$ 0.008 & 0.012 $\pm$  0.011 \\
ORB-SLAM (short) & F & 0.064 $\pm$ 0.141 & 0.064 $\pm$ 0.130 \\
\hline
Zhou \etal \cite{zhou2017unsupervised} & R & 0.021 $\pm$ 0.017 & 0.020 $\pm$ 0.015 \\
Zhou \etal \cite{zhou2017unsupervised} \dag& R & 0.665 $\pm$ 0.150 & 0.467 $\pm$ 0.202 \\
% GeoNet \cite{yin2018geonet} ResNet& R & 0.012 $\pm$0.007 & 0.012 $\pm$ 0.009 \\
% GeoNet \cite{yin2018geonet} VGG & R & 0.012 $\pm$0.007 & 0.012 $\pm$ 0.009 \\
% \hline
% Mahjourian \etal \cite{mahjourian2018unsupervised} & R+ICP &0.013 $\pm$ 0.010 & 0.012 $\pm$ 0.011\\
\hline
% \midrule
DSO (full) & D & 0.065 $\pm$ 0.059 & 0.047 $\pm$ 0.043  \\
% Wang et al. \cite{wang2018learning}* & D& 0.063 $\pm$0.126 & 0.085 $\pm$ 0.115 \\
Wang \etal  \cite{wang2018learning} & R+D & 0.045 $\pm$0.108 & 0.033 $\pm$ 0.074 \\

\midrule
Our method (3-frame) & D & 0.017 $\pm$ 0.017 &  0.011 $\pm$ 0.010 \\ 
Our method (5-frame) & D & 0.025 $\pm$ 0.020 & 0.016 $\pm$ 0.014\\ 
% \hline
\bottomrule
\end{tabular}
\end{center}{}
\caption{Absolute Trajectory Error (ATE) on the KITTI odometry dataset. The results of other baselines are taken from \cite{zhou2017unsupervised,yin2018geonet,wang2018learning}. For the type column, F  denotes the feature-based method, R denotes the CNN regression method and D denotes the direct method. The DSO results are taken from \cite{klodt2018supervising}. Zhou\dag  denotes the pose estimation results that take both forward and backward as inputs in the training process. For our method, we evaluate the integrated poses of 3-frame and 5-frame snippets.}
\label{tab:kitti_pose}
\end{table}

\section{Conclusion}
% We construct the multiscale feature and depth pyramids for pose estimation. 
We propose an end-to-end unsupervised learning framework for monocular visual odometry with the direct method. Our scheme can tightly incorporate features at different levels for pose estimation by constructing the multiscale feature and depth pyramids. The scale factor can be constrained with our method; furthermore, our scheme does not need the forward motion constraints. This work reveals the capability of the deep learning method to contribute to the direct visual odometry pipeline. It can be extended by incorporating it with other widely applied modules, such as marginalization and keyframe selection, to construct a complete SLAM system, which is out of the scope of this paper. 

% \textcolor{red}{
% We propose the jointly unsupervised learning framework GeoNet, and demonstrate the advantages of exploiting ge- ometric relationships over different previously “isolated” tasks. Our unsupervised nature profoundly reveals the capa- bility of neural networks in capturing both high level cues and feature correspondences for geometry reasoning. The impressive results compared to other baselines including the supervised ones indicate possibility of learning these low level vision tasks without costly collected groundtruth data.
% }

There are some interesting directions for follow-up works. In particular, our method can benefit from the incorporation of nonrigid object removal using  optical flow prediction (e.g., \cite{vijayanarasimhan2017sfm,yin2018geonet}) and tracking. % Optical flow prediction for  could further boost the performances.

%-------------------------------------------------------------------------

\clearpage

%-------------------------------------------------------------------------

{\small
\bibliographystyle{ieee}
\bibliography{egbib}
}

\end{document}